\documentclass[letterpaper, 10 pt, conference]{ieeeconf}  

\IEEEoverridecommandlockouts                              

\usepackage{etoolbox} 

\usepackage{moreverb,url}
\usepackage{amsmath} 
\usepackage{amssymb} 
\usepackage{todonotes}
\let\origtodo\todo
\newcommand{\basetodo}{\origtodo}

\renewcommand{\todo}[1]{\basetodo[inline]{#1}}
\usepackage[most]{tcolorbox}    
\usepackage{subcaption} 
\usepackage{dsfont} 
\usepackage{upgreek} 
\usepackage{accents} 
\usepackage{textcomp}
\usepackage{makecell} 
\usepackage{diagbox}
\usepackage{hhline}
\usepackage{hyperref}
\usepackage{stfloats} 

\newcommand{\gravitycoriolis}{\mathbf{h}} 
\newcommand{\jangle}{q}
\newcommand{\jangles}{\mathbf{\jangle}} 
\newcommand{\jvelocity}{\dot{\jangle}}
\newcommand{\jvelocities}{\mathbf{\jvelocity}} 
\newcommand{\jacceleration}{\ddot{\jangle}}
\newcommand{\jaccelerations}{\mathbf{\jacceleration}} 

\newcommand{\position}{x}
\newcommand{\eeposition}{\mathbf{\position}} 
\newcommand{\eevelocity}{\mathbf{\dot{\position}}} 
\newcommand{\eeacceleration}{\mathbf{\ddot{\position}}} 

\newcommand{\numberOfConstraint}{\mathcal{K}}
\newcommand{\numberOfObjectContacts}{B} 
\newcommand{\numberOfJoints}{D}
\newcommand{\humanWrench}{\mathbf{F}} 
\newcommand{\graspmatrix}{\mathbf{G}} 
\newcommand{\jacobian}{\mathbf{J}} 
\newcommand{\jacobianDot}{\mathbf{\dot{J}}} 
\newcommand{\numberOfFootContacts}{L}
\newcommand{\inertia}{\mathbf{M}} 
\newcommand{\projection}{\mathbf{P}} 
\newcommand{\rotationMatrix}{\mathbf{R}} 
\newcommand{\selectionMatrix}{\mathbf{S}} 
\newcommand{\BMatrix}{\mathbf{B}}

\newcommand{\wrench}{\boldsymbol{\lambda}}
\renewcommand{\pi}{\uppi}
\newcommand{\eeWrench}{\mathbf{F}_c} 
\newcommand{\torque}{\tau}
\newcommand{\torqueVector}{\boldsymbol{\torque}} 

\newcommand{\tsInertia}{\boldsymbol{\Lambda}} 

\newcommand{\nullVector}{\mathbf{0}} 
\newcommand{\nullMatrix}{\mathbf{0}} 
\newcommand{\zeroMatrix}{\mathbf{0}} 
\newcommand{\identityMatrix}{\mathbf{I}} 
\newcommand{\realNumbers}{\mathds{R}} 
\newcommand{\skewMatrix}{\mathbb{S}}

\newcommand{\norm}[1]{\left\lVert#1\right\rVert}

\definecolor{gray}{rgb}{0.5,0.5,0.5}

\newcommand{\refsec}[1]{Sec.~\ref{#1}}

\newcommand{\stiffness}{\boldsymbol{K}_{p}}
\newcommand{\damping}{\boldsymbol{K}_{d}}

\title{\LARGE \bf
Shared Object Manipulation with a Team of Collaborative Quadrupeds
}

\author{Shengzhi Wang$^{1, *}$, Niels Dehio$^{2}$, Xuanqi Zeng$^{1}$, Xian Yang$^{1}$, Lingwei Zhang$^{1}$, \\ Yun-Hui Liu$^{1}$, and K. W. Samuel Au$^{1}$
\thanks{$^{1}$Department of Mechanical and Automation Engineering at The Chinese University of Hong Kong, Hong Kong, China. *: Corresponding author}%
\thanks{$^{2}$Technology and Innovation Center (TIC), KUKA Deutschland GmbH.}%
}

\newtheorem{remark}{Remark}

\begin{document}

\maketitle
\thispagestyle{empty}
\pagestyle{empty}


\begin{abstract}
Utilizing teams of multiple robots is advantageous for handling bulky objects. 
%
Many related works focus on multi-manipulator systems,
which are limited by workspace constraints.
%
In this paper, we extend a classical hybrid motion-force controller  
to a team of legged manipulator systems,
enabling collaborative loco-manipulation of rigid objects with a force-closed grasp. 
%
Our novel approach allows the robots to flexibly coordinate their movements, 
achieving efficient and stable object co-manipulation and transport,
validated through extensive simulations and real-world experiments.
\end{abstract}

\section{INTRODUCTION}
Controlling a team of collaborative robots has become a hot research topic,
as these teams enable the handling of large and bulky objects that would be impossible for a single robot.
Scenarios are often inspired by applications in logistics, where boxes need to be grabbed~\cite{WangYuquan2023}, rotated, transported, tossed~\cite{Bombile2023} 
and/or caught in flight~\cite{Salehian2016,Yan2024}. 

Robot teams aiming at shared object manipulation and transportation 
may be composed of fixed-base manipulators, i.e.,
dual-arm~\cite{Bonitz1996,Caccavale2008,Dehio2022a} 
and multi-arm systems~\cite{Hayati1986,Peng2018,Dehio2022b},
however, these systems have a limited workspace.
To address this limitation,
multiple floating-base robots equipped with manipulators have been combined, 
such as mobile manipulators~\cite{Marino2017,Marino2018}, 
humanoids~\cite{Vaillant2017,Bouyarmane2019}, 
drones~\cite{ParraVega2013} 
and underwater vehicles~\cite{Simetti2017}.

\begin{figure}[t!]
            \centering
            \includegraphics[width=0.495\textwidth]{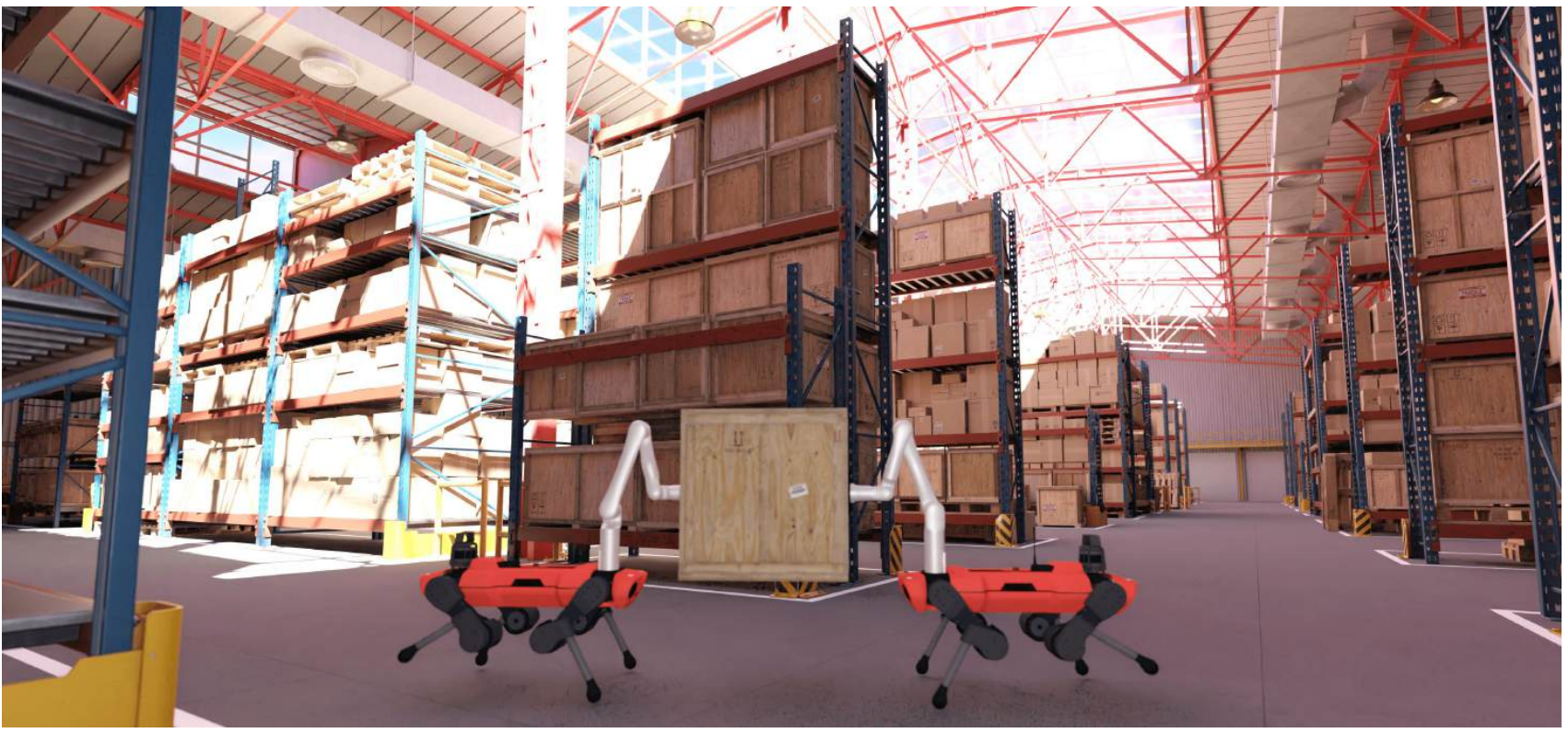}
            \caption{An object manipulation scenario with a robot team.}
            \label{fig:application_example}
\end{figure}

To our best knowledge, 
there has been little exploration of using a team of legged manipulators for joint object manipulation requiring a force-closed grasp,
such as transporting objects (see Fig.~\ref{fig:application_example}).
We address this gap with a novel torque-control scheme for collaborative loco-manipulation of rigid, bulky objects with multiple legged manipulators.
We formulate a Cartesian impedance for the object and simultaneously resolve the remaining redundancy in the robot team.
Our simulations and real-world experiments demonstrate efficient and stable joint co-manipulation and transport of an object.

\begin{table*}[!b]
\caption{Summary of Related Works on Contact-Consistent Motion Generation with PIDC.
}
\centering
\footnotesize
\setlength{\tabcolsep}{5.1pt}
\begin{tabular}{|c|c|c|c|c|}
\hline
 & \thead{without contact force control,\\without inequality constraints} 
 & \thead{with contact force control,\\without inequality constraints} 
 & \thead{without contact force control,\\with inequality constraints} 
 & \thead{with contact force control,\\with inequality constraints} 
 \\
 \hline
\thead{no underactuation} 
 & trivial 
 & \cite{Aghili2005}
 & --
 & \cite{Lin2018}
 \\
 \hline 
\thead{underactuation\\treated numerically}
 & --
 & --
 & \cite{Aghili2016}
 & \cite{Righetti2013}
 \\
 \hline
\thead{underactuation\\treated in closed-form}
 & \cite{righetti2011inverse, Mistry2012}
 & \cite{Dehio2018}
 & --
 & \textbf{Proposed}
 \\
 \hline
\end{tabular}
\label{tab:summary}
\end{table*}

\subsection{Related Work on Hybrid Motion-Force Control}

Shared control of an object is challenging due to the physical coupling and exchange of contact forces, necessitating both a force-closed grasp and simultaneous motion control.
Within this work, we are particularly interested in a hybrid motion-force control concept called
\emph{Projected Inverse Dynamics Control} (PIDC),
originally introduced in~\cite{Aghili2005} 
for a single fully actuated stationary robot. 
Extensions to the underactuated case 
were presented in \cite{Righetti2013,Aghili2016} based on numeric optimization methods.
In contrast, \cite{Dehio2018,righetti2011inverse,Mistry2012} found a closed-form solution to that problem,
with \cite{Dehio2018} generalizing \cite{righetti2011inverse, Mistry2012} by explicitly also controlling contact forces.
%
%
%
Considering unilateral contacts, additional friction cone constraints must be considered. 
These constraints, expressed as inequalities, can be explicitly handled using quadratic programming to minimize power consumption, as demonstrated for manipulators in~\cite{Lin2018}.

In this paper, we formulate a coherent control scheme that combines
the closed-form PIDC approach for underactuated robots~\cite{Dehio2018}, 
with a contact force optimization for fully actuated robots~\cite{Lin2018}.
With appropriate substitutions, 
our generic formulation simplifies to~\cite{Dehio2018} or~\cite{Lin2018}, also indicated in table~\ref{tab:summary}.
%
%
Hence, 
our main contribution is a low-level hybrid motion-force control scheme for torque-controlled underactuated multi-robot systems,
enabling collaborative loco-manipulation tasks.
For the first time, we demonstrate a force-closed grasp with a team of legged manipulators.



\section{FUNDAMENTALS}
The constrained floating-base multi-robot dynamics, subjected to external forces and expressed in Lagrangian form, are modeled as a rigid multi-body system with $\numberOfJoints$ joints, including both virtual and passive joints
\begin{equation}
 \label{eq:js_dynamics}
 \inertia \jaccelerations + \gravitycoriolis = \selectionMatrix \torqueVector + \jacobian_{c}^{T} \wrench + \jacobian_{x}^{T} \humanWrench 
   \text{ ,}
\end{equation}
where
$ \jangles $, $ \jvelocities $, $ \jaccelerations \in \realNumbers^{\numberOfJoints} $ are generalized coordinates, $\torqueVector  \in \realNumbers^{\numberOfJoints} $ is the vector of joint torques,
$\selectionMatrix \in \realNumbers^{\numberOfJoints \times \numberOfJoints}$ is a diagonal matrix selecting actuated joints, 
$ \inertia \in \realNumbers^{\numberOfJoints \times \numberOfJoints} $ is the inertia matrix,
and $ \gravitycoriolis \in \realNumbers^{\numberOfJoints} $ compensates for gravity and Coriolis effects.
For readability, we omit the dependency notation on $ \jangles $ and $ \jvelocities $. Rigid contact constraints, assuming zero Cartesian velocities and accelerations, are modeled with negligible elasticity
\begin{equation}
 \label{eq:contact_constraint}
 \jacobian_{c} \, \jvelocities = \nullVector 
 \text{  and  } 
 \jacobian_{c} \, \jaccelerations + \jacobianDot_{c} \, \jvelocities = \nullVector
 \text{ ,}
\end{equation}
where $ \jacobian_{c} \in \realNumbers^{\numberOfConstraint \times \numberOfJoints} $ 
describes the constrained Jacobian 
associated with $\numberOfConstraint$ constraints
and $ \wrench \in \realNumbers^{\numberOfConstraint}$ denotes the vertical concatenation of all contact wrenches. 
The external disturbances $\humanWrench \in \realNumbers^{6}$ from human interaction or environment are projected onto the joint-space via the associated Jacobian $\jacobian_{x} \in \realNumbers^{6 \times \numberOfJoints}$.
The control goal is to generate torque commands $\torqueVector$ 
such that $\wrench$ satisfy contact constraints (represented as inequalities) while executing a desired motion.


\subsection{Projected Inverse Dynamics Control}
The PIDC approach~\cite{Aghili2005} ensures dynamic consistency with contact constraints 
by addressing them at the highest priority level within a strict task hierarchy 
utilizing an orthogonal projection matrix $\projection$ and $\identityMatrix - \projection$, 
projecting~\eqref{eq:js_dynamics} into two subspaces
\begin{equation}
 \label{eq:aghili_21}
    \projection \left( \inertia \jaccelerations + \gravitycoriolis \right) = \projection \selectionMatrix \torqueVector + \projection \jacobian_{x}^{T} \humanWrench \text{,}
\end{equation}
\begin{equation}
 \label{eq:aghili_31_5}
 (\identityMatrix - \projection) 
 \left( \inertia \jaccelerations + \gravitycoriolis \right) = (\identityMatrix - \projection) \selectionMatrix \torqueVector + \jacobian_c^T \wrench + (\identityMatrix - \projection) \jacobian_{x}^{T} \humanWrench
 \text{,}
\end{equation}
where $ \projection \in \realNumbers^{\numberOfJoints \times \numberOfJoints} = \identityMatrix - \jacobian_{c}^T {\left( \jacobian_{c}^+ \right)}^T$ projects vectors onto the null space of the constraint (\textit{motion space}), such that $\projection = \projection^2 = \projection^T$, $\projection \jacobian_c^T = \boldsymbol{0}$, and $\projection \jvelocities = \jvelocities$ (equivalently, $(\identityMatrix - \projection) \jvelocities = \boldsymbol{0}$) for all $\jvelocities \in \mathcal{N}\left(\jacobian_c\right)$, and $\jacobian_{c}^+$ is the Moore-Penrose inverse. Then $\identityMatrix - \projection$ represents the complementary projection into $\mathcal{N}^{\perp}(\jacobian_c)$ (\textit{constraint space}). 

\subsection{Inferring Joint Accelerations and Contact Wrenches}
To predict $\wrench$, 
joint accelerations $\jaccelerations$ must be computed. Since $\projection$ might be rank deficient, $\jaccelerations$ cannot be computed directly by pre-multiplying $(\projection \inertia)^{-1}$ from~\eqref{eq:aghili_21}. Instead, solve for $\jaccelerations$ utilizing the property $(\identityMatrix-\projection)\jvelocities = \boldsymbol{0}$ and its derivative $(\identityMatrix-\projection) \jaccelerations = \dot{\projection} \jvelocities$, by substituting the latter equation into~\eqref{eq:aghili_21}
\begin{equation}
\label{eq:aghili_25}
 \jaccelerations=\inertia_{c}^{-1} (\projection \selectionMatrix \torqueVector - \projection \gravitycoriolis +\dot{\projection} \jvelocities + \projection \jacobian_{x}^{T} \humanWrench ), 
\end{equation}
with $\inertia_{c} = \projection \inertia + \identityMatrix - \projection$. Inserting~\eqref{eq:aghili_25} into~\eqref{eq:aghili_31_5}, the constraint wrenches are computed,
where $\bar{\inertia} = \inertia \inertia_{c}^{-1}$ is the so-called \textit{constraint inertia matrix} which is always invertible~\cite{Aghili2005}
\begin{equation}
\begin{aligned}
        \label{eq:constraint_force}
         \wrench=(\jacobian_{c}^{T})^+\Big[&(\identityMatrix - \projection)[\bar{\inertia}(\projection \selectionMatrix \torqueVector - \projection \gravitycoriolis + \dot{\projection} \jvelocities) + \gravitycoriolis] \\ 
         &- (\identityMatrix - \projection) \selectionMatrix \torqueVector + (\identityMatrix-\projection)(\bar{\inertia} \projection - \identityMatrix)\jacobian_{x}^{T} \humanWrench \Big].
\end{aligned}
\end{equation}
This equation is used in~\refsec{subsec:numeric_optimization_of_contact_forces} to find the relation between the decision variables and the contact wrenches $\wrench$.

\subsection{Task-Space Cartesian Impedance Control}
\label{sec:intro_impedance_control}

A Cartesian impedance controller is applied in the motion space following~\cite{Hogan1985,Ott2008,Xin2020,WangShengzhi2023}.
The control law writes
\begin{equation}
        \label{eq:control_law}
        \humanWrench_{d, x}=\gravitycoriolis_{c}+\tsInertia_{c} \eeacceleration_{d}-\boldsymbol{K}_{d} \dot{\boldsymbol{e}}-\boldsymbol{K}_{p} \boldsymbol{e}, 
\end{equation}
where $\tsInertia_{c} = (\jacobian_x \inertia_c^{-1} \projection \jacobian_x^T)^{-1}$ is the \textit{operational space inertia matrix}, $\gravitycoriolis_c = \tsInertia_{c} \jacobian_{x} \inertia_{c}^{-1}(\projection \gravitycoriolis-\dot{\projection} \jvelocities)-\tsInertia_{c} \dot{\jacobian}_{x} \jvelocities$ is the \textit{operational space non-linear effect}, $\boldsymbol{K}_{d}$ and $\boldsymbol{K}_{p}$ are the desired task-space damping and stiffness matrix gains, and $\boldsymbol{e} = \eeposition - \eeposition_d$ denotes the pose error of the end-effector between the current and desired one. Although $\tsInertia_{c}$ is robot configuration dependent, the stability of the closed-loop dynamics between $\boldsymbol{e}$ and $\humanWrench$ is still guaranteed~\cite{Ott2008}. 

Eventually, the implementation at the joint level yields:
\begin{equation}
\label{eqn:joint_space_tau_M}
    \torqueVector_{\text{M}} = \jacobian_x ^T \humanWrench_{d, x}\,.
\end{equation}
\begin{remark}
\label{remark:intro_impedance_control}
The subscript `x' in~\eqref{eqn:joint_space_tau_M} represents a particular task space of interest. 
\end{remark}

\subsection{Projected Inverse Dynamics for Underactuated Robots}

Here, we briefly reiterate the control law from~\cite{Dehio2018} solving the underactuation constraint
through closed-form projection
\begin{equation} 
 \label{eq:ua_solution}
 \!\!\!
 \torqueVector = 
   {\left[  \projection \selectionMatrix  \right]}^+ 
   \projection \torqueVector_{\text{M}} 
 + \! \BMatrix\!
   \left( \identityMatrix \text{$\,$\textminus$\,$} \projection \right) \torqueVector_{\text{C}} 
   \text{,}
\end{equation}
where
\begin{equation}
\label{eq:bmatrix}
\BMatrix = { \identityMatrix \text{$\,$\textminus$\,$} 
   {\left[ \left( \identityMatrix \text{$\,$\textminus$\,$} \selectionMatrix \right) \!
   \left( \identityMatrix \text{$\,$\textminus$\,$} \projection \right) \right]}^+ }.
\end{equation}

\subsection{Force-Closed Grasp}
\label{subsec:force_closed_grasp}
Consider multi-fingered manipulation of a rigid object with $\numberOfObjectContacts$ contacts.
The \textit{partial grasp matrix} for the $i^{th}$~fingertip 
$\graspmatrix_i \in \realNumbers^{6 \times 6}$ 
maps contact twists to the object twist~\cite{a_mathematical_introduction_to_robotic_manipulation}
\begin{equation}
 \graspmatrix_{i} = 
 \begin{bmatrix}
  \rotationMatrix_i             & \zeroMatrix      \\
  \skewMatrix(\boldsymbol{r}_i) & \rotationMatrix_i 
 \end{bmatrix},
\end{equation}
with the rotation matrix~$\rotationMatrix_i$ of the $i^{th}$ contact frame, 
the distance vector~$\boldsymbol{r}_i$ from the object center of mass to the $i^{th}$ contact frame,
and the skew-symmetric matrix for the cross~product~$\skewMatrix(\boldsymbol{r}) \in \realNumbers^{3 \times 3}$.
The \textit{complete grasp matrix} 
is formed by horizontally concatenating all $\numberOfObjectContacts$ partial grasp matrices into~$\graspmatrix \in \realNumbers^{6 \times 6 \, \numberOfObjectContacts}$,
representing the relative transformations at all contact locations
\begin{equation}
 \label{eq:graspmatrix}
 \graspmatrix = \left[ \graspmatrix_{1}, ... , \graspmatrix_{\numberOfObjectContacts} \right].
\end{equation}

The nullspace projection of the complete grasp matrix, given by  
$\identityMatrix - \graspmatrix^T {\left( \graspmatrix^+ \right)}^T \in \realNumbers^{6 \, \numberOfObjectContacts \times 6 \, \numberOfObjectContacts}$, projects any vector onto the grasp matrix's nullspace. 
The resulting contact wrench, referred to as the internal force, generates no net wrench, expressed as $\graspmatrix \wrench_{ee} = \nullVector \in \realNumbers^{6}$, where $\wrench_{ee} \in \realNumbers^{6 \, \numberOfObjectContacts}$. This property is used in~\refsec{sec:contact_constraints_in_collaborative_control}.
The classical concept of multi-fingered manipulation also applies to multiple robot arms jointly manipulating a rigid object~\cite{Dehio2022b}.
We follow this approach and refer to a fingertip in contact as the \emph{hand} of a legged manipulator.

\section{PROPOSED METHOD}
\label{sec:methodology}
\begin{figure}[t!]
            \centering
            \includegraphics[width=0.495\textwidth]{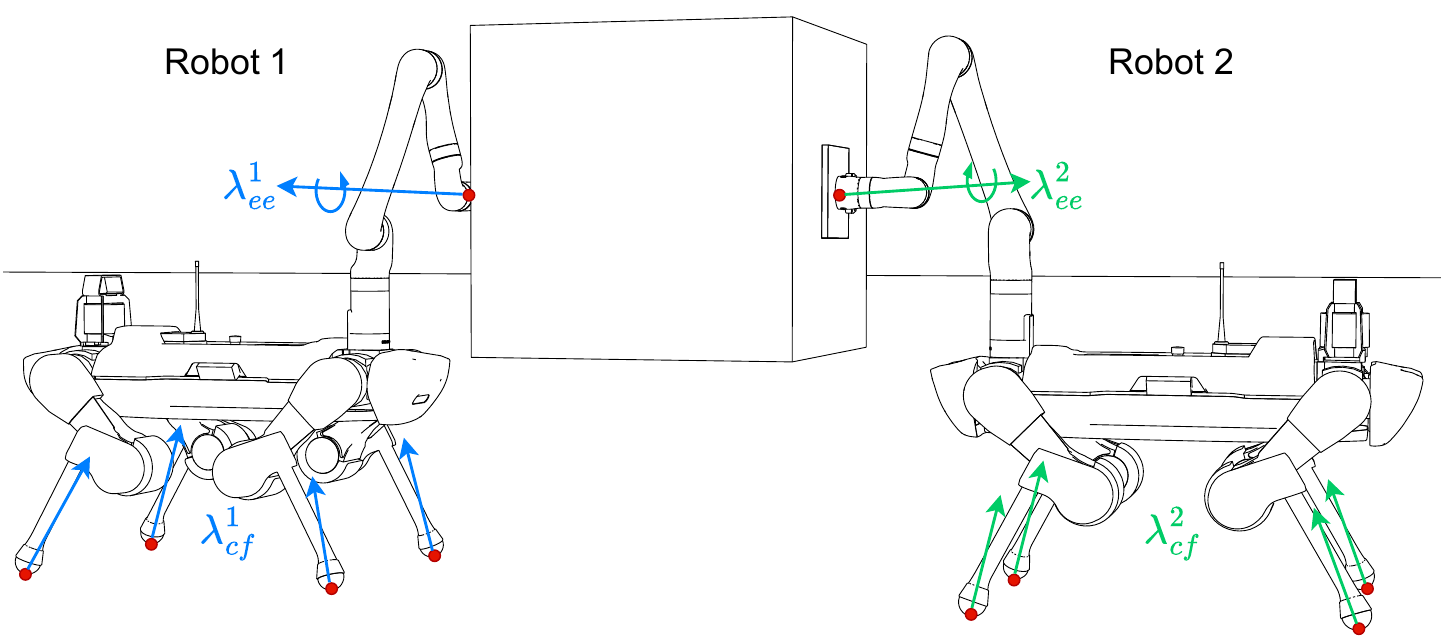}
            \caption{Graphical representation of two collaborative quadrupeds with manipulators attached to the trunk. The foot contact force vector $\lambda^i_{cf}$ stacks the ground reaction forces on each foot in contact, whereas the contact wrench $\lambda^i_{ee} \in \realNumbers^{6}$ describes the interaction between the box and the hand.}
            \label{fig:collaborative_manipulation_sketch}
\end{figure}

We propose a coherent control scheme for collaborative object manipulation with a team of underactuated multi-legged manipulators (see Fig.~\ref{fig:collaborative_manipulation_sketch}),
generalizing~\cite{Dehio2018} and~\cite{Lin2018}.
The scheme realizes a 6-Degree of Freedom (DoF) Cartesian impedance both at the object and the robot torsos and feet, 
while optimizing constraint forces at the contact points.

\subsection{Contact Constraints in Collaborative Control}
\label{sec:contact_constraints_in_collaborative_control}
We consider a multi-robot system composed of underactuated legged manipulators 
having $\numberOfFootContacts$ feet in contact with the static environment, 
and a manipulated rigid object with $\numberOfObjectContacts$ contact locations,
e.g., robot hands\footnote{Later on, we also consider contact between object and environment.}. 
This system is modeled by~\eqref{eq:js_dynamics} with $\numberOfJoints$-DoF. 
%
As all legged manipulators operate independently, the inertia matrix $ \inertia \in \realNumbers^{\numberOfJoints \times \numberOfJoints}$ is block-diagonal
and $\gravitycoriolis \in \realNumbers^{\numberOfJoints}$ aggregates their contributions. 
We focus on point-like feet (3 constraints in each contact)
and flat surface-like object contacts (6 constraints each),
but our formulation can easily adapt to other contact conditions, like humanoids with flat feet or spherical fingertips.

To construct the constraint Jacobian $\jacobian_{c} \in \realNumbers^{(3 \numberOfFootContacts + 6 \numberOfObjectContacts)  \times \numberOfJoints}$, 
we differentiate between the combined contact feet Jacobian $\jacobian_{cf} \in \realNumbers^{3 \numberOfFootContacts \times \numberOfJoints}$ and a combined hand Jacobian $\jacobian_{ee} \in \realNumbers^{6 \numberOfObjectContacts \times \numberOfJoints} $. 
The $\jacobian_{cf}$ consists of $\numberOfFootContacts$ contact foot Jacobians $\jacobian_{cf, i} \in \realNumbers^{3 \times \numberOfJoints}$,
whereas $\jacobian_{ee}$ is formed from $\numberOfObjectContacts$ hand Jacobians $\jacobian_i \in \realNumbers^{6 \times \numberOfJoints}$. 
During collaborative manipulation, constraints must ensure a rigid grasp while allowing unrestricted motion generation. 
Contact wrenches are controlled to permit only the internal wrench. 
Thus, following~\refsec{subsec:force_closed_grasp}, the constraint hand Jacobian for the rigid grasp $\jacobian_{c, ee} \in \realNumbers^{6 \numberOfObjectContacts \times \numberOfJoints}$ is
\begin{equation}
 \label{eq:jacobian_cee}
 \jacobian_{c, ee} = \left( \identityMatrix - \graspmatrix^T {\left( \graspmatrix^+ \right)}^T \right) \jacobian_{ee}\,.
\end{equation}
Note that $\jacobian_{c, ee}$ constrains only the relative motion between hands, 
ensuring a virtual linkage.
Absolute motion is of course permitted as long as it does not affect the relative transformation between the hands. Additionally, $\jacobian_{cf}$ constrains the absolute motion of all feet in contact.

\begin{remark}
\label{remark:for_single_robot_press_wall}
\eqref{eq:graspmatrix} and~\eqref{eq:jacobian_cee} can also model a scenario with a rigid object pressed against a wall, where the contact location with the wall defines a partial grasp matrix and the associated Jacobian contains zeros only.
\end{remark}

Finally, the constraint Jacobian $\jacobian_c$ is constructed from the contact foot Jacobian $\jacobian_{cf}$
and the rigid grasp Jacobian $\jacobian_{c, ee}$
\begin{equation}
 \label{eq:jacobian_c}
 \jacobian_{c} = [\jacobian_{cf}^T, \jacobian_{c,ee}^T]^T.
\end{equation}
The Jacobian $\jacobian_{o} \in \realNumbers^{6 \times \numberOfJoints}$ for the object center of mass is
\begin{equation}
 \label{eq:obj_jacobian}
 \jacobian_{o} = {\left( \graspmatrix^+ \right)}^T \jacobian_{ee} \,.
\end{equation}

\subsection{Cartesian Impedance Controller}
\label{subsec:Cartesian Impedance Controller}
Sec.~\ref{sec:intro_impedance_control} introduces the Cartesian impedance feedback law for motion control. As Remark~\ref{remark:intro_impedance_control} states, the desired motion of the quadrupeds' torsos and swing feet can be achieved by replacing the $\jacobian_x$ with the relevant Jacobians. For the carried object, the impedance controller uses the Jacobian $\jacobian_o$. The motion control torque $\torqueVector_{\text{M}}$ is then the superposition of all impedance controllers mapped back to joint space. 

The current object pose $\eeposition_{o}$ is inferred from the assumption of a rigid grasp with a constant relative transformation between the hands and the object's center, eliminating the need for external sensors like vision systems. Slip can be detected by evaluating $\jacobian_{c} \, \jvelocities \neq \nullVector$ to update the grasp matrix~\cite{Dehio2022b}.

\subsection{Numeric Optimization of Contact Forces}
\label{subsec:numeric_optimization_of_contact_forces}
Projections enable the closed-form solution of various equality constraints, but inequality constraints for physical consistency, like friction cone and unilateral constraints, cannot be solved in closed-form. 
Inspired by~\cite{Dehio2018,Lin2018}, we propose combining the extended PIDC formalism~\eqref{eq:ua_solution} with quadratic programming
to seek reference torques that satisfy these inequality constraints related to contact properties.

In this paper, 
the subscripts $()_f$ and $()_m$ represent force and moment, respectively. The contact wrench is defined as 
$\wrench = \left[\wrench_{f,x}, \wrench_{f,y}, \wrench_{f,z}, \wrench_{m,x}, \wrench_{m,y}, \wrench_{m,z} \right]^T$,
where $\wrench_{f,z}$ is the normal force, 
$\wrench_{f,x}, \wrench_{f,y}$ are tangential forces,
and $\wrench_{m,x}, \wrench_{m,y}, \wrench_{m,z}$
are moments along each axis. 

\subsubsection{Objective Function}
To reduce overall power consumption, we determine the lowest actuator torques required to maintain all contacts as
\begin{equation}
\label{eq:start_objective_function}
 \min_{\torqueVector} \, \torqueVector^T \torqueVector 
 \text{ .}
\end{equation}
Substituting $\torqueVector = 
   {\left[  \projection \selectionMatrix  \right]}^+ 
   \projection \torqueVector_{\text{M}} 
 + \! \BMatrix\!
   \left( \identityMatrix \text{$\,$\textminus$\,$} \projection \right) \torqueVector_{\text{C}}$ 
\eqref{eq:ua_solution}
yields:
\begin{equation}
\label{eq:wrench_optim0}
\min_{ \torqueVector = 
   {\left[  \projection \selectionMatrix  \right]}^+ 
   \projection \torqueVector_{\text{M}} 
 + \! \BMatrix\!
   \left( \identityMatrix \text{$\,$\textminus$\,$} \projection \right) \torqueVector_{\text{C}} } \left\{ \,
\begin{IEEEeqnarraybox}[][c]{l?s}
\IEEEstrut
 \left[ {\left[  \projection \selectionMatrix  \right]}^+ 
   \projection \torqueVector_{\text{M}} \right]^T \! {\left[  \projection \selectionMatrix  \right]}^+ 
   \projection \torqueVector_{\text{M}} 
 \\
 + \left[ {\left[  \projection \selectionMatrix  \right]}^+ 
   \projection \torqueVector_{\text{M}} \right]^T \!\BMatrix \left( \identityMatrix - \projection \right) \torqueVector_{\text{C}}
 \\
 + \left[\BMatrix \left( \identityMatrix - \projection \right) \torqueVector_{\text{C}} \right]^T {\left[  \projection \selectionMatrix  \right]}^+ 
   \projection \torqueVector_{\text{M}}
 \\
 + \left[\BMatrix \left( \identityMatrix - \projection \right) \torqueVector_{\text{C}} \right]^T \! \BMatrix \left( \identityMatrix - \projection \right) \torqueVector_{\text{C}}
\IEEEstrut
\text{,}
\end{IEEEeqnarraybox}
\right.
\end{equation}
where $\torqueVector_{\text{C}}$ denotes the contact wrenches to be determined from the optimization problem. 
During optimization, the term $\left[ {\left[  \projection \selectionMatrix  \right]}^+ 
   \projection \torqueVector_{\text{M}} \right]^T \! {\left[  \projection \selectionMatrix  \right]}^+ 
   \projection \torqueVector_{\text{M}}$ can be eliminated as it is constant with respect to $\torqueVector_{\text{C}}$. Furthermore, due to orthogonal subspaces, $\left[ {\left[  \projection \selectionMatrix  \right]}^+ 
   \projection \torqueVector_{\text{M}} \right]^T \!\!\BMatrix \left( \identityMatrix - \projection \right) \torqueVector_{\text{C}}$ and $\left[\BMatrix \left( \identityMatrix - \projection \right) \torqueVector_{\text{C}} \right]^T \! {\left[  \projection \selectionMatrix  \right]}^+ 
   \projection \torqueVector_{\text{M}}$ equal zero. Thus, it suffices to minimize $\left[\BMatrix \left( \identityMatrix - \projection \right) \torqueVector_{\text{C}} \right]^T \! \BMatrix \left( \identityMatrix - \projection \right) \torqueVector_{\text{C}}$ only in~\eqref{eq:wrench_optim0}
\begin{IEEEeqnarray}{rCl}
 \label{eq:wrench_optim1}
 & &\min_{\BMatrix\left( \identityMatrix - \projection \right) \torqueVector_{\text{C}}} \, \left[\BMatrix \left( \identityMatrix - \projection \right) \torqueVector_{\text{C}} \right]^T \! \BMatrix \left( \identityMatrix - \projection \right) \torqueVector_{\text{C}}
 \text{ .}
\end{IEEEeqnarray}

Let $\eeWrench \in \realNumbers^{3 \numberOfFootContacts + 6 \numberOfObjectContacts}$ be the unknown wrenches at the hands and feet corresponding to the input torque, satisfying
$\left( \identityMatrix \text{$\,$\textminus$\,$} \projection \right) \torqueVector_{\text{C}} 
\! = \! ( \identityMatrix \text{$\,$\textminus$\,$} \projection ) \, \jacobian_{c}^{T} \eeWrench
\! = \! \jacobian_{c}^{T} \! \eeWrench $. 
Since $\jacobian_{c}^{T}$ and $\projection$ 
remain constant during optimization, and $( \identityMatrix - \projection ) \jacobian_{c}^{T} = \jacobian_{c}^{T}$, the objective function can be rewritten in terms of~$\eeWrench$ through
\begin{IEEEeqnarray}{C}
 \label{eq:wrench_optim2}
 \min_{\eeWrench} \, \eeWrench^T \jacobian_{c} ( \identityMatrix - \projection )^T \BMatrix^T \BMatrix ( \identityMatrix - \projection ) \, \jacobian_{c}^{T} \eeWrench
 \nonumber \\
 = \min_{\eeWrench} \, \eeWrench^T \jacobian_{c} \BMatrix^T \BMatrix  \mspace{1mu} \jacobian_{c}^{T} \eeWrench
 \text{ .}
\end{IEEEeqnarray}
\begin{remark}
\eqref{eq:wrench_optim2} generalizes the formulation from~\cite{Lin2018} by introducing the new term $\BMatrix^T \BMatrix$,
taking into account underactuation~\eqref{eq:bmatrix} as suggested in~\cite{Dehio2018}.
Our novel expression simplifies to the solution presented in~\cite{Lin2018} when considering a fully-actuated system,
i.e., $\BMatrix = \identityMatrix$ with $\selectionMatrix = \identityMatrix$.
This shows the generic nature, 
combining previous works~\cite{Dehio2018} and~\cite{Lin2018}.
\end{remark}

\subsubsection{Inequality Constraints for Physical Consistency}
\label{sec:inequiality_constraints}
In the following, four types of inequality constraints are considered:

\paragraph{Unilateral and Friction Cone Constraints}
\label{para:Unilateral and Friction Cone Constraints}
Let $\wrench_{f, i} \in \realNumbers^{3}$ be the contact force at the $i$-th contact, with $\boldsymbol{n}_{x, i}$, $\boldsymbol{n}_{y, i}$, $\boldsymbol{n}_{z, i} \in \realNumbers^{3}$ as the heading, lateral, and normal vectors of the contact surface, and $\mu_i$ as the friction coefficient. The unilateral and friction cone constraints are
\begin{equation}
\label{eqn::unilateral_and_friction_cone_constraint}
   \underbrace{\left[\begin{array}{c}-\infty \\ -\infty \\ 0 \\ 0 \\ 0 \end{array}\right]}_{\underline{\boldsymbol{d}}} \! \leq \! \underbrace{\left[\begin{array}{c}(\boldsymbol{n}_{x, i} - \mu_i \boldsymbol{n}_{z, i})^T \\ (\boldsymbol{n}_{y, i} - \mu_i \boldsymbol{n}_{z, i})^T \\ (\boldsymbol{n}_{x, i} + \mu_i \boldsymbol{n}_{z, i})^T \\ (\boldsymbol{n}_{y, i} + \mu_i \boldsymbol{n}_{z, i})^T \\ \boldsymbol{n}_{z, i}^T\end{array}\right]}_{\boldsymbol{C}_i} \boldsymbol{\lambda}_{f, i} \! \leq \! \underbrace{\left[\begin{array}{c}0 \\ 0 \\ +\infty \\ +\infty \\ +\infty \end{array}\right]}_{\overline{\boldsymbol{d}}},
\end{equation}
where $\underline{\boldsymbol{d}}, \overline{\boldsymbol{d}} \in \realNumbers^{5}$ are the lower/upper bound vector, $\boldsymbol{C}_i \in \realNumbers^{5 \times3}$ is the constraint matrix for the $i$-th contact. The first four rows of~\eqref{eqn::unilateral_and_friction_cone_constraint} describe the approximated friction cone model (the friction pyramid~\cite{Trinkle1997}), while the last row encodes the unilateral constraint. 

\paragraph{Moment Constraints}
\label{para:Moment Constraints}
 Let $\wrench_{i} \in \realNumbers^{6}$ be the contact wrench at the $i$-th contact. To prevent rolling at contact points, we assume the contact patch and surface friction are sufficient to generate both frictional force and moment, so constraints are imposed on the torsional moment~\cite{Buss1996} and shear moment~\cite{Bonitz1996} as:
\begin{equation}
\label{eqn::moment_constraint}
   \underbrace{\left[\begin{array}{c}\!0\!\! \\ \!0\!\! \\ \!0\!\! \\ \!0\!\! \\ \!0\!\! \\ \!0\!\!\end{array}\right]}_{\underline{\boldsymbol{d}}_m} \!\! \leq \! \underbrace{ \begin{bmatrix}
  0 & 0 & \xi & 0 & 0 & -1 \\
  0 & 0 & \xi & 0 & 0 & 1 \\
  0 & 0 & \delta_{x} & -1 & 0 & 0 \\
  0 & 0 & \delta_{x} & 1 & 0 & 0 \\
  0 & 0 & \delta_{y} & 0 & -1 & 0 \\
  0 & 0 & \delta_{y} & 0 & 1 & 0 
 \end{bmatrix}}_{\boldsymbol{C}_{m, i}} \boldsymbol{\lambda}_{i} \leq \!\! \underbrace{\left[\begin{array}{c}\!+\infty\!\! \\ \!+\infty\!\! \\ \!+\infty\!\! \\ \!+\infty\!\! \\ \!+\infty\!\! \\ \!+\infty\!\! \end{array}\right]}_{\overline{\boldsymbol{d}}_m},
\end{equation}
where $\xi$ is the torsional friction coefficient, and $\delta_{x}, \delta_{y}$ are the distances from the contact patch center to its edges in the X and Y directions, respectively (assuming a rectangular contact patch). 

\paragraph{Torque Constraints}
To ensure that the final torque commands meet actuator saturation limits, we add
\begin{equation}
    \label{eq:original_torque_limit}
    \torqueVector_{min} \leq \selectionMatrix_{ns}^T \torqueVector \leq \torqueVector_{max}, 
\end{equation}
where $\torqueVector_{min}, \torqueVector_{max} \in \realNumbers^{\mathcal{L}}$ are the lower/upper bounds of all $\mathcal{L}$ actuated joints, and $\selectionMatrix_{ns} \in \realNumbers^{D\times \mathcal{L}}$ is a non-square selection matrix that eliminates torque values of the virtual joints from $\torqueVector$. By replacing $\left( \identityMatrix \text{$\,$\textminus$\,$} \projection \right) \torqueVector_{\text{C}} 
\! = \! \jacobian_{c}^{T} \! \eeWrench$ into~\eqref{eq:ua_solution} and then into~\eqref{eq:original_torque_limit}, the torque limit constraints w.r.t the decision variable $\humanWrench_c$ are expressed as
\begin{equation}
    \label{eq:real_torque_limit}
    \underline{\boldsymbol{\torqueVector}} \leq \Xi \, \humanWrench_c \leq \overline{\boldsymbol{\torqueVector}}, 
\end{equation}
with $\Xi = \selectionMatrix_{ns}^T \BMatrix \jacobian_c^T$,
and vectors $\underline{\boldsymbol{\torqueVector}} = \torqueVector_{min} - \selectionMatrix_{ns}^T {\left[  \projection \selectionMatrix  \right]}^+ 
   \projection \torqueVector_{\text{M}}$ 
and $\overline{\boldsymbol{\torqueVector}} = \torqueVector_{max} - \selectionMatrix_{ns}^T {\left[  \projection \selectionMatrix  \right]}^+ 
   \projection \torqueVector_{\text{M}}$. 

Note that in \textit{a)} and \textit{b)}, the actual contact wrenches $\wrench$ differ from the commanded wrenches $\eeWrench$. Their relationship can be derived from~\eqref{eq:constraint_force}, which contains terms $\projection \selectionMatrix \torqueVector$ and $(\identityMatrix - \projection) \selectionMatrix \torqueVector$ that have not yet been expanded. 
By substituting $\left( \identityMatrix \text{$\,$\textminus$\,$} \projection \right) \torqueVector_{\text{C}} 
\! = \! \jacobian_{c}^{T} \! \eeWrench$ into~\eqref{eq:ua_solution} and then inserting the resulting control law $\torqueVector$, we can expand these terms as
\begin{equation}
\label{eq:PStau_original}
    \projection \selectionMatrix \torqueVector = \projection \selectionMatrix {\left[  \projection \selectionMatrix  \right]}^+ 
   \projection \torqueVector_{\text{M}} 
 + \! \projection \selectionMatrix\BMatrix
   \jacobian_{c}^{T} \! \eeWrench,
\end{equation}
\begin{equation}
\label{eq:IminusPStau_original}
    (\identityMatrix - \projection) \selectionMatrix \torqueVector = (\identityMatrix - \projection) \selectionMatrix {\left[  \projection \selectionMatrix  \right]}^+ 
   \projection \torqueVector_{\text{M}} 
 + \! (\identityMatrix - \projection) \selectionMatrix\BMatrix
   \jacobian_{c}^{T} \! \eeWrench.
\end{equation}
Insert the expansions of $\projection \selectionMatrix \torqueVector$ and $(\identityMatrix - \projection) \selectionMatrix \torqueVector$ from~\eqref{eq:PStau_original} and~\eqref{eq:IminusPStau_original} back into~\eqref{eq:constraint_force}, the relation between $\wrench$ and $\eeWrench$ yields
\begin{IEEEeqnarray}{rl}
        \label{eq:relation_between_lambda_and_Fc}
         \wrench=&(\jacobian_{c}^{T})^+\Big[(\identityMatrix \!-\! \projection)[\bar{\inertia}(\projection\selectionMatrix {\left[  \projection \selectionMatrix  \right]}^+ \projection
   \torqueVector_{\text{M}} \!+ \! \projection\selectionMatrix\BMatrix\jacobian_{c}^{T}\eeWrench 
\\ 
         &\!-\! \projection \gravitycoriolis\!+\! \dot{\projection} \jvelocities) \! + \! \gravitycoriolis] \!-\! (\identityMatrix \!-\! \projection) \selectionMatrix{\left[  \projection \selectionMatrix \right]}^+\projection 
   \torqueVector_{\text{M}} 
\nonumber \\
 &\!-\! (\identityMatrix \!-\! \projection) \selectionMatrix \BMatrix \jacobian_{c}^{T} \! \eeWrench \!+ \!(\identityMatrix \!-\! \projection)(\bar{\inertia} \projection \!-\! \identityMatrix)\jacobian_{x}^{T} \humanWrench \Big]=\boldsymbol{\rho} \eeWrench \!+\! \boldsymbol{\eta},
 \nonumber
\end{IEEEeqnarray}
where 
\begin{equation*}
    \boldsymbol{\rho} = (\jacobian_{c}^{T})^+ (\identityMatrix \!-\! \projection) (\bar{\inertia} \projection \!-\! \identityMatrix)\selectionMatrix \BMatrix \jacobian_{c}^{T},
\end{equation*}
\begin{equation*}
    \boldsymbol{\eta} = (\jacobian_{c}^{T})^+ (\identityMatrix \!-\! \projection) [\boldsymbol{\varepsilon} \torqueVector_{\text{M}} + \bar{\inertia}(\!-\! \projection \gravitycoriolis + \dot{\projection} \jvelocities) + \gravitycoriolis + (\bar{\inertia} \projection \!-\! \identityMatrix)\jacobian_{x}^{T} \humanWrench],
\end{equation*}
\begin{equation*}
    \boldsymbol{\varepsilon} = (\bar{\inertia} \projection \!-\! \identityMatrix) \selectionMatrix{\left[  \projection \selectionMatrix  \right]}^+\projection.
\end{equation*}

Note again, that~\eqref{eq:relation_between_lambda_and_Fc} generalizes~\cite{Lin2018} for underactuated systems
by considering matrix $\BMatrix$~\eqref{eq:bmatrix} derived in~\cite{Dehio2018}.
Substituting~\eqref{eq:relation_between_lambda_and_Fc} into~\eqref{eqn::unilateral_and_friction_cone_constraint} and~\eqref{eqn::moment_constraint} introduces the decision variables $\eeWrench$ into these constraints.

\begin{table}[b]
\centering
\caption{Control Parameters for Simulations}
\label{table:control_parameter_simulations}
\begin{tabular}{|c|c|} 
\hline
\multicolumn{2}{|c|}{\begin{tabular}[c]{@{}c@{}}Tasks for simultaneous torso and object motion\end{tabular}}  \\ 
\hline
Torso Stiffness Gains   & diag(4000, 4000, 4000, 3000, 3000, 3000)                                                                           \\ 
\hline
Torso Damping Gains     & diag(200, 200, 200, 125, 125, 125)                                                                                 \\ 
\hline
Object Stiffness Gains & diag(1500, 1500, 1500, 400, 400, 400)                                                                              \\ 
\hline
Object Damping Gains   & diag(150, 150, 150, 140, 140, 140)                                                                                 \\ 
\hhline{|==|}
\multicolumn{2}{|c|}{Task for Quadruped Walking}                                                                                            \\ 
\hline
Torso Stiffness Gains   & diag(1200, 1200, 1200, 600, 600, 600)                                                                              \\ 
\hline
Torso Damping Gains     & diag(100, 100, 100, 50, 50, 50)                                                                                    \\ 
\hline
Feet Stiffness Gains & diag(3500, 3500, 3500)                                                                                 \\
\hline
Feet Damping Gains & diag(20, 20, 20)                                                                                 \\
\hline
Object Stiffness Gains & diag(500, 500, 500, 400, 400, 400)                                                                                 \\ 
\hline
Object Damping Gains   & diag(50, 50, 50, 40, 40, 40)                                                                                       \\
\hline
\end{tabular}
\end{table}

\begin{figure*}[t!]
            \centering
            \includegraphics[width=\textwidth]{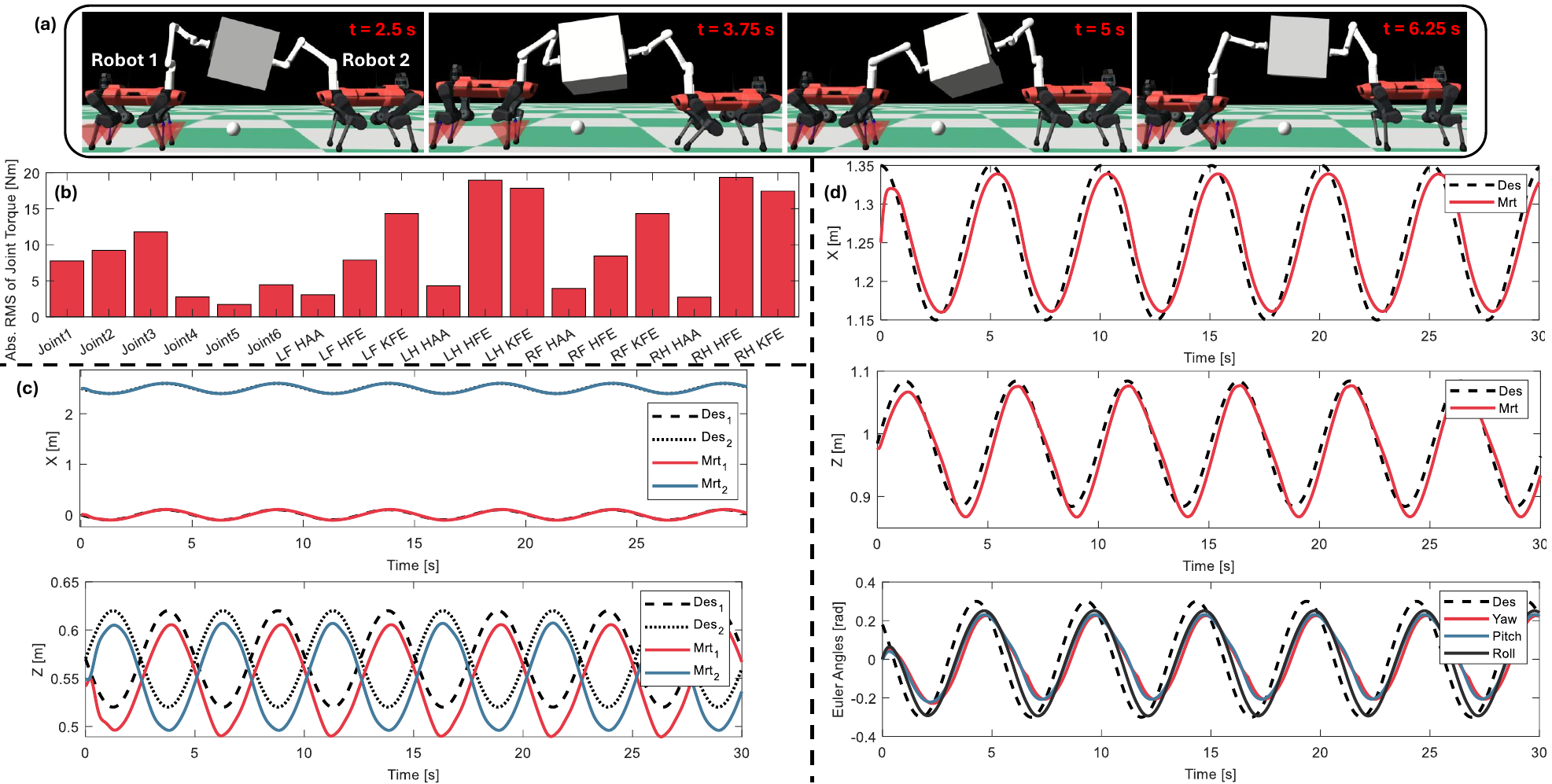}
            \caption{Simultaneous torso and object motion: (a) Simulation snapshots; (b) Absolute RMS values of commanded joint torques for robot $1$ (joint $1$ - $6$ belong to the manipulator).
LF: left front leg; RF: right front leg; LH: left hind leg; RH: right hind leg; HAA: hip abduction/adduction; HFE: hip flexion/extension; KFE: knee flexion/extension; (c) Quadruped torso motions; (d) Object motion.
`Des' is the desired value,
`Mrt' is the measured value,
with subscripts indicating the robot.}
            \label{fig:results_hybrid_motion_task}
\end{figure*}

\begin{figure*}[t!]
            \centering
            \includegraphics[width=\textwidth]{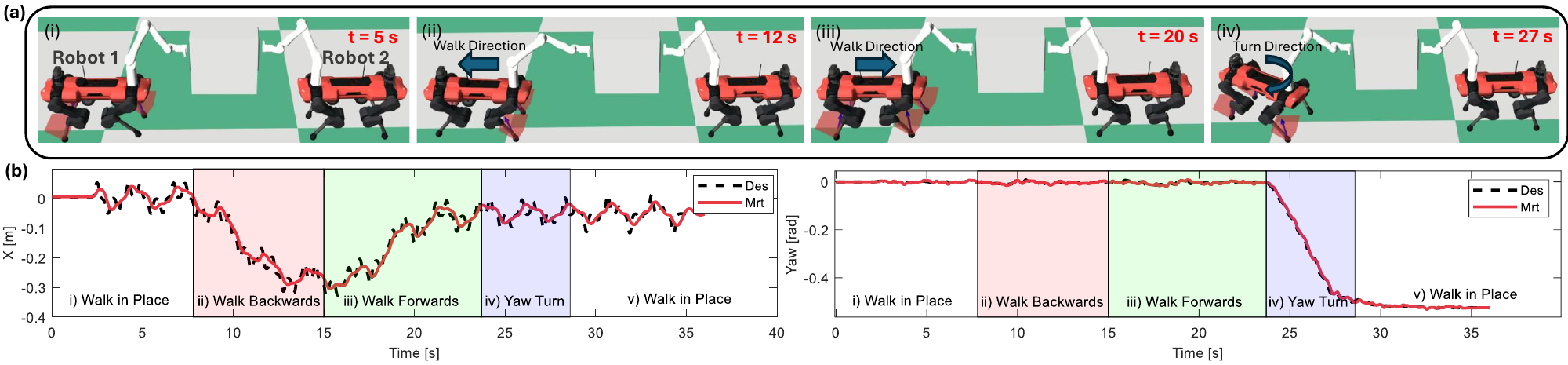}
            \caption{Quadruped walking: (a) Simulation snapshots show (i) walk in place, (ii) walk backwards, (iii) walk forwards, and (iv) Yaw turn; (b) Torso motion of robot~$1$. `Des' is the desired value, `Mrt' is the measured value.}
            \label{fig:results_of_quadruped_walking_task}
            \vspace{-0.45cm}
\end{figure*}

\section{SIMULATIONS}

The proposed control scheme is validated in the physics engine RaiSim~\cite{raisim}.
A cubic box ($0.6 \,\text{m}$ length,
$1 \,
\text{kg}$ mass) is co-manipulated by two legged manipulator systems through a force-closed grasp,
each consisting of a 6-DoF Kinova Gen~3 robot arm attached to the trunk of the 12-DoF quadruped ANYmal~C.
The robot hand is designed as a rectangular flat contact surface similar to~\cite{Dehio2018,Lin2018}.
Following prior work on collaborative object manipulation, we assume the controller has knowledge of the object's shape, desired contact points, and surface normals.
We provide preplanned Cartesian reference trajectories in the object,
quadruped torso,
and feet frames.
It is legitimate to assume that these reference trajectories are consistent and compatible for the entire robot system,
although discussing a feasible planner is beyond the scope of this paper.
Note that contact wrench profiles for the robot hands and feet, 
or joint position references do not need to be provided, 
as these are automatically inferred by the controller.
Also, the motion of the robot hands is not explicitly controlled.
We focus on two scenarios to demonstrate collaborative (loco-)manipulation: 
$(i)$ simultaneous torso and object motion, 
and $(ii)$~quadruped walking.
Table~\ref{table:control_parameter_simulations} lists the control parameters.

\subsection{Simultaneous Torso and Object Motion}
\label{sec:simultaneous_torso_and_object_motion}
We provide sinusoidal trajectories along the X and Z axes for the torso,
and in X, Z, Roll, Pitch, and Yaw direction for the object,
as depicted in Fig.~\ref{fig:results_hybrid_motion_task}~\hyperref[fig:results_hybrid_motion_task]{(a)}.
Fig.~\ref{fig:results_hybrid_motion_task}~\hyperref[fig:results_hybrid_motion_task]{(b)}
presents the absolute root mean square (RMS) values of the commanded torque at each joint of robot~$1$,
skipping robot~$2$ due to symmetry effects.
The hind leg joints,
particularly the second (HFE) and third (KFE),
exhibit larger torque distributions.
Fig.~\ref{fig:results_hybrid_motion_task}~\hyperref[fig:results_hybrid_motion_task]{(c)} and~\hyperref[fig:results_hybrid_motion_task]{(d)} 
demonstrate accurate tracking performance for the quadruped torsos and the manipulated object.

\subsection{Quadruped Walking}
This scenario involves robot~$1$ walking backwards for $0.27\,\text{m}$ utilizing a classical static walk gait pattern, 
then returning to its initial position,
and finally turning Yaw for the torso by $-30^{\circ}$,
see Fig.~\ref{fig:results_of_quadruped_walking_task}~\hyperref[fig:results_hybrid_motion_task]{(a)}.
Meanwhile,
robot~$2$ torso and the object track a constant pose.
Our controller successfully tracks the reference locomotion trajectories, 
cf. Fig.~\ref{fig:results_of_quadruped_walking_task}~\hyperref[fig:results_hybrid_motion_task]{(b)}.

\subsection{Controller Comparison}
\begin{figure}[t!]
            \centering
            \includegraphics[width=.4\textwidth]{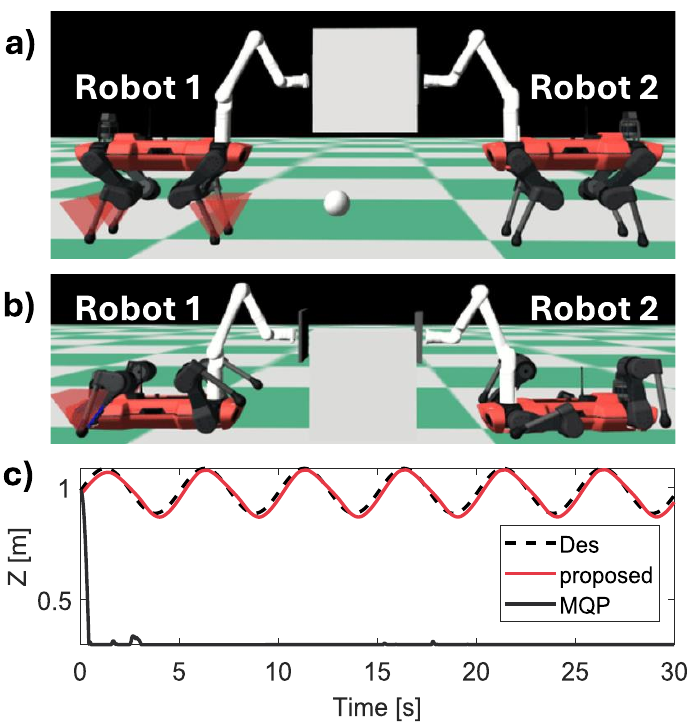}
            \caption{Controllers comparison: (a) Maintaining a constant torso and object pose using MQP; (b) Failure of MQP when performing simultaneous torso and object motion (as described in Sec. \ref{sec:simultaneous_torso_and_object_motion}); (c) Object's Z motion under different controllers.}
            \label{fig:controller_comparison}
            \vspace{-0.5cm}
\end{figure}

We compare our controller with the multirobot QP (MQP) controller proposed in~\cite{Bouyarmane2019} for simultaneous torso and object motion, as described in Sec \ref{sec:simultaneous_torso_and_object_motion}:
    \begin{subequations}
    \begin{equation}
         \min_{ \jaccelerations, \wrench, \torqueVector } \hspace{0.2cm}
         \norm{
         \begin{bmatrix}
         \jacobian_x & \!\!\!\!  \nullMatrix & \!\!\!\!  \nullMatrix
         \end{bmatrix} \!\!
         \begin{bmatrix}
         \jaccelerations \\ \wrench \\ \torqueVector
         \end{bmatrix} 
         \! - \!
         \left(
         \eeacceleration_{\text{ref}}
         - \jacobianDot_x \jvelocities_{\text{mrt}}
         \right)
         }^2, 
    \end{equation}
    \begin{equation}
    \begin{split}
    \text{s. t. \,\,\,} &\text{Dynamics, Contact Kinematics, Torque,}\\
    &\text{Joint Acceleration and Velocity,}\\
    &\text{and Friction Cone Constraints,}
    \end{split}
    \end{equation}
    \end{subequations}
    with 
    \begin{equation}
         \eeacceleration_{\text{ref}} =  \eeacceleration_{\text{des}}
         + \stiffness (\eeposition_{\text{des}} - \eeposition_{\text{mrt}}) 
         + \damping (\eevelocity_{\text{des}} - \eevelocity_{\text{mrt}}), 
    \end{equation}
    where $\eeposition$ denotes the multidimensional tasks, $()_{\text{mrt}}$ denotes the measured values, $()_{\text{des}}$ denotes the desired values, and $\jacobian_x$ is the Jacobian matrix of the multidimensional tasks. Here, the multidimensional tasks include the $6$-DoF task space of the carried box, the torsos of quadruped robots, and the $3$-DoF task space of the swing legs. The MQP can achieve a constant pose tracking task 
 for the object and the torsos, see Fig.~\ref{fig:controller_comparison}~\hyperref[fig:controller_comparison]{(a)}. However, it fails to achieve the simultaneous torso and object motion compared with our proposed controller. see Fig.~\ref{fig:controller_comparison}~\hyperref[fig:controller_comparison]{(b)} and ~\hyperref[fig:controller_comparison]{(c)}.

\begin{table}[b]
\centering
\caption{Control Parameters for Hardware Experiments}
\label{table:control_parameter_hardware}
\begin{tabular}{|c|c|} 
\hline
Torso Stiffness Gains   & diag(2000, 2000, 2000, 375, 375, 375)                                                                           \\ 
\hline
Torso Damping Gains     & diag(50, 50, 50, 15, 15, 15)                                                                                 \\ 
\hline
Object Stiffness Gains & diag(250, 250, 250, 50, 50, 50)                                                                              \\ 
\hline
Object Damping Gains   & diag(35, 35, 35, 7.5, 7.5, 7.5)                                                                                 \\
\hline
\end{tabular}
\end{table}

\section{HARDWARE VERIFICATION}

\begin{figure}[t!]
            \centering
            \includegraphics[width=.485\textwidth]{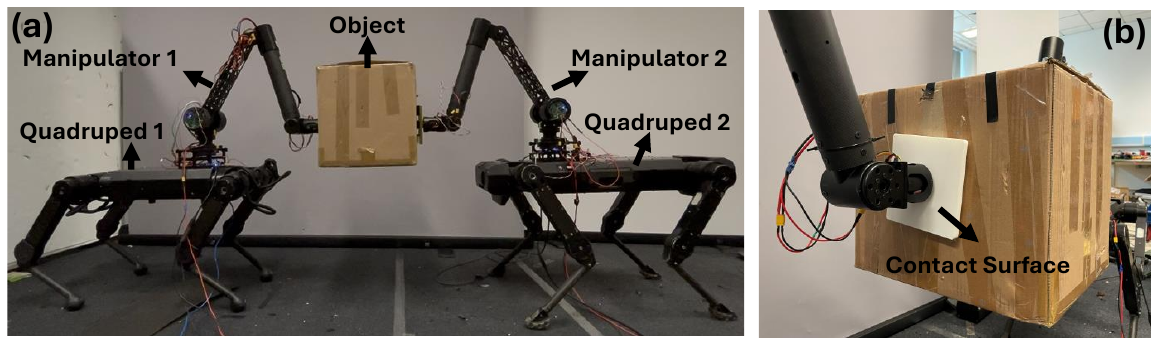}
            \caption{Hardware setup: (a) Front view; (b) Side view.}
            \label{fig:hardware_setup}
\end{figure}

For real-world validation, we use two self-built torque-controlled quadrupeds with actuated 12-DoF each, Sirius-Belt (Quadruped $1$) and Sirius-Diff (Quadruped $2$), 
both are equipped with 6-DoF manipulators (Sirius-Arm).
Robot dimensions are $0.78 \times 0.37 \times 0.54$ m (length $\times$ width $\times$ height) when fully standing.
The challenge is to co-manipulate a $1\, \text{kg}$ box with two flat contact surface hands.  
The hardware setup is illustrated in Fig.~\ref{fig:hardware_setup}, 
and the control parameters for all experiments are listed in Table~\ref{table:control_parameter_hardware}. 
Torso and box pose are estimated using a linear Kalman filter~\cite{flayols2017experimental}. 




\subsection{Object Circular Motion}
We command the box to follow a circular trajectory in the X-Z plane while keeping its orientation and the quadrupeds' torso poses fixed. Figure \ref{fig:circular_motion} illustrates the tracking performance, demonstrating that the controller successfully maintains a force-closed grasp while accurately following the reference trajectories.

\subsection{Object Orientation Motion}
\begin{figure}[t!]
            \centering
            \includegraphics[width=.39\textwidth]{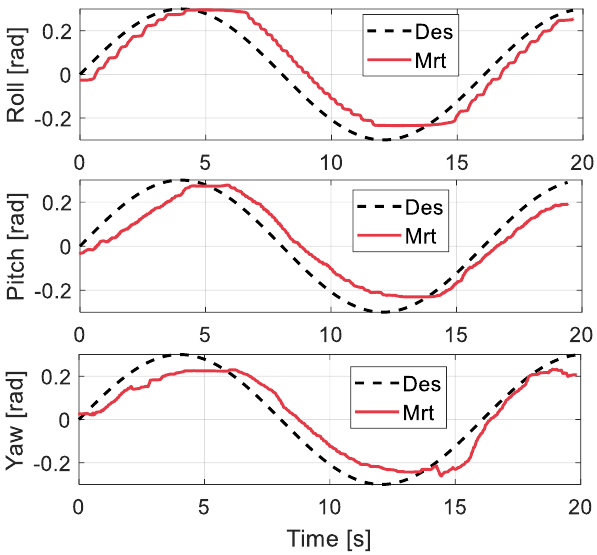}
            \caption{Object orientational motion in Roll, Pitch, and Yaw, respectively. `Des' is the desired value, `Mrt' is the measured value.}
            \label{fig:roll_pitch_yaw_motion}
            \vspace{-0.4cm}
\end{figure}

We command sinusoidal movements for the box in Roll, Pitch, and Yaw, respectively, while keeping its position and the quadrupeds' torso poses fixed. Successful tracking performances in all three axes are shown in Fig. \ref{fig:roll_pitch_yaw_motion}.
With the proposed controller, the two robots jointly rotate the object as commanded.

\subsection{Cartesian Impedance Characteristics}
\begin{figure*}[t!]
            \centering
            \includegraphics[width=\textwidth]{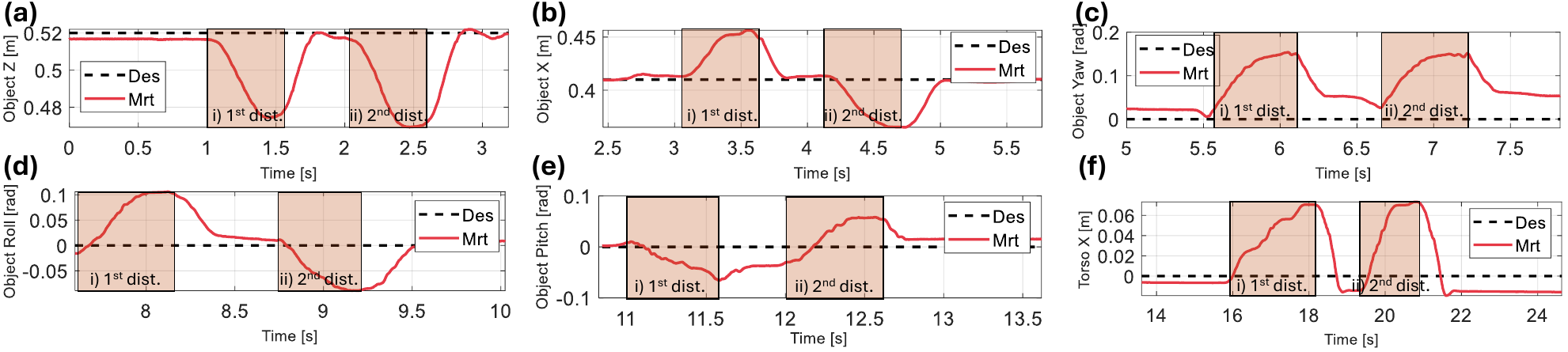}
            \caption{Impedance behavior of the object and torso: (a) - (e) Object motion in Z, X, Yaw, Roll, and Pitch direction, respectively; (f) Torso motion of quadruped $1$.
`Des' is the desired value,
`Mrt' is the measured value,
the orange areas in each subfigure are the first and second perturbation phases.}
            \label{fig:perturbation_tests}
\end{figure*}

Figure \ref{fig:perturbation_tests} illustrates the dual-robot system response to external disturbances, highlighting its Cartesian impedance characteristics. The robots initially hold the object in a fixed, desired pose. 
External perturbations are then introduced in a controlled manner. 
First, the object is perturbed sequentially along the Z and X axes, 
as well as in the rotational Yaw, Roll, and Pitch directions over time. 
Next, an additional perturbation is applied to the torso of quadruped~$1$ along the X~direction. 
The controller complies with the external forces and actively restores both the object and the robots to their desired configurations once the external force subsides.
Successful recovery demonstrates that the grasp remains force-closed and stable throughout the perturbations.

\begin{figure}[t!]
            \centering
            \includegraphics[width=.39\textwidth]{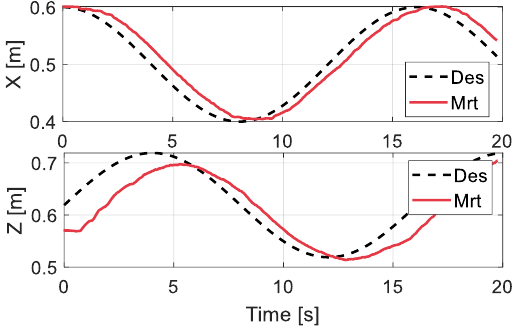}
            \caption{Object circular motion in X and Z direction. `Des' is the desired value, `Mrt' is the measured value.}
            \label{fig:circular_motion}
\end{figure}

\section{DISCUSSION}
Extensive simulation and hardware experiments confirm that the proposed controller effectively maintains a force-closed grasp on the object throughout collaborative loco-manipulation, with no signs of slippage or instability. Smooth tracking results validate its ability to regulate contact forces while in motion, ensuring a secure and stable grasp.
Cartesian impedance control enables the system to compensate for external forces. When subjected to disturbances, it maintains grasp stability, ensuring robustness in real-world conditions.

In Fig. \ref{fig:circular_motion} and \ref{fig:roll_pitch_yaw_motion}, a slight tracking delay is observed, primarily due to motor friction. 
Friction, particularly in the 2nd and 3rd motors in each arm, increases under high loads as the synchronous belt compresses the bearings and transmission structures, causing instability and leading to inconsistent speeds with a fixed PID parameter set. 

Compared to the baseline MQP controller, the proposed approach demonstrates superior tracking accuracy and stability in dynamic scenarios, preventing robot instability and object misalignment. This extends the feasible task space for multi-quadruped manipulation beyond prior methods.

In summary, the proposed controller achieves stable, coordinated loco-manipulation of a shared object, laying the groundwork for more advanced collaborative tasks in the future.

\section{CONCLUSIONS and FUTURE WORK}
This paper presents a novel hybrid motion-force control scheme for collaborative loco-manipulation 
of rigid, bulky objects using pure frictional contacts in floating-base multi-legged manipulator systems. 
Generalizing the PIDC approach, 
the unconstrained (motion) controller tracks several trajectories in task spaces associated with the object, robot torsos, and feet,
imposing Cartesian impedance. 
The constrained component enforces contact constraints by numerically optimizing the contact forces.
The control scheme is evaluated on torque-controlled legged manipulators through various simulations and real-world experiments, 
demonstrating its feasibility and effectiveness for multi-robot teams.

Future work aims at integrating a centralized motion planning strategy similar to~\cite{culbertson2021decentralized}, 
creating a comprehensive collaborative loco-manipulation framework. 

\section*{ACKNOWLEDGMENT}
The authors thank all members of CUHKLRL for their discussions and help, and especially thank Dr. Linzhu Yue for his generous collaboration.


\bibliographystyle{./IEEEtran} 
\bibliography{ref}

\end{document}